\documentclass[10pt,journal,letterpaper]{IEEEtran}
\usepackage[T1]{fontenc}
\usepackage{times}
\usepackage{amsmath,amssymb}
\usepackage{graphicx}
\usepackage{trimclip} 
\usepackage{booktabs,array,tabularx}
\usepackage{algorithm}
\usepackage{algpseudocode}
\usepackage{cite}
\usepackage{microtype}
\usepackage{balance}
\usepackage{placeins} 
\usepackage{capt-of}
\usepackage{float}
\usepackage{xcolor}
\definecolor{projectblue}{RGB}{52,112,169}

\usepackage[hidelinks]{hyperref}
\hypersetup{pdftitle={BioAgent: Self-Directed Skill Acquisition and Capability Growth for Long-Horizon Laboratory Protocols},pdfauthor={}}
\graphicspath{{figures/}}
\newcolumntype{Y}{>{\raggedright\arraybackslash}X}
\newcommand{\K}{\mathcal{K}}
\newcommand{\B}{\mathcal{B}}
\newcommand{\T}{\mathcal{T}}
\newcommand{\D}{\mathcal{D}}
\newcommand{\Hacc}{\mathcal{H}_{\mathrm{acc}}}
\newcommand{\SR}{\mathrm{SR}}
\newcommand{\best}[1]{\textbf{#1}}

\usepackage{color}

\begin{document}
\title{BioAgent: Self-Directed Skill Acquisition and Capability Growth for Long-Horizon Laboratory Protocols}
\author{
Gaojing Zhang$^{1,2,*}$,
Changwen Chen$^{1,*}$,
Zichen Luo$^{1,*}$,
Jialiang Li$^{1}$,
Yuhan Wang$^{1}$,
Yifei Xie$^{1}$,
Wenzhao Lian$^{1,{\dagger}}$
\\[3pt]
{\small
$^{1}$Shanghai Jiao Tong University
\qquad
$^{2}$University of Sussex
}
\\[2pt]
{\footnotesize
$^{*}$Equal Contribution.
$^{\dagger}$Corresponding author
}
}

\IEEEaftertitletext{%
  \vspace{-2.25\baselineskip}
  \begin{minipage}{\textwidth}
    \centering
    \includegraphics[width=\textwidth]{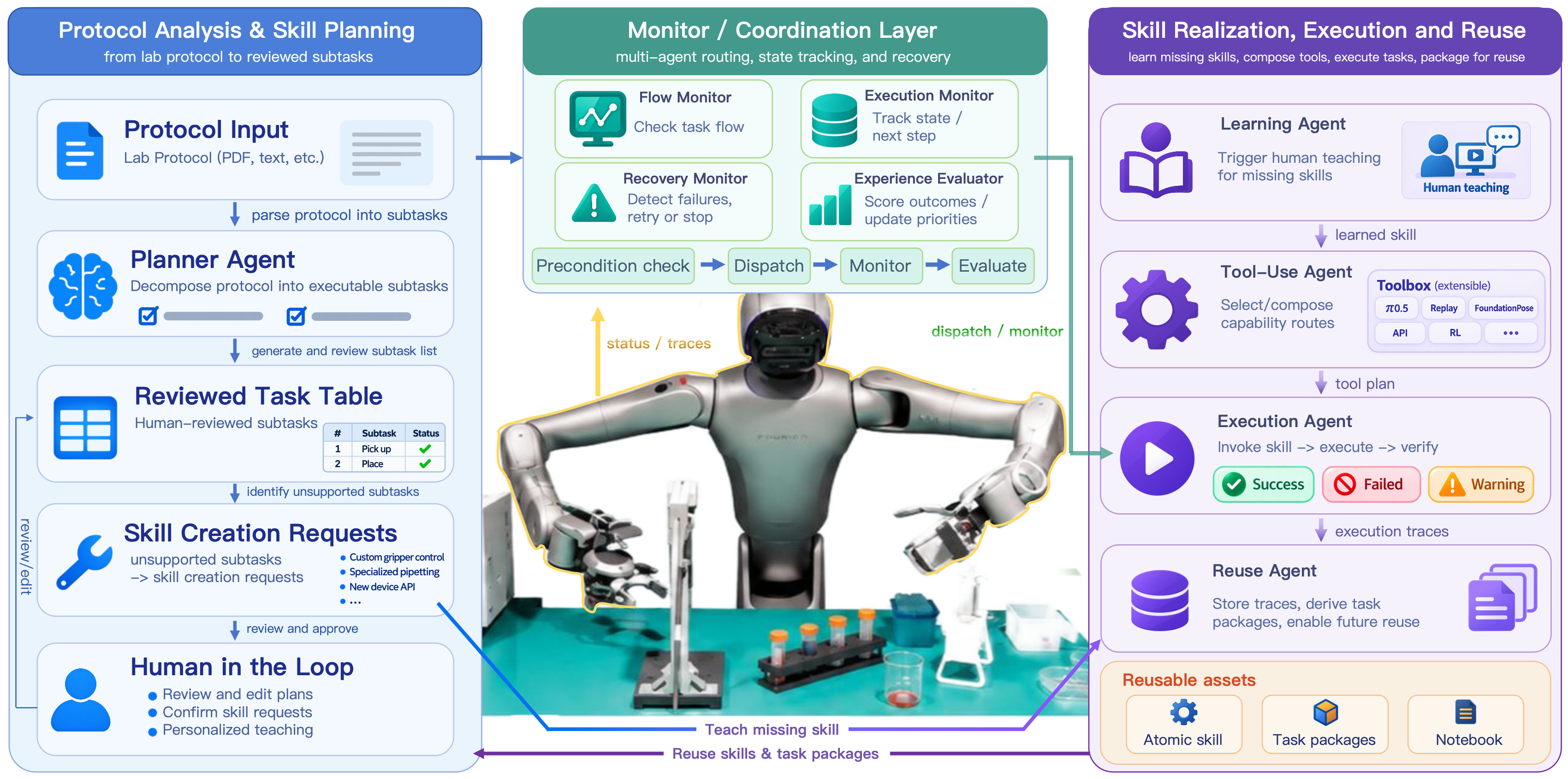}
    \captionof{figure}{BioAgent overview. Human-reviewed protocol requirements guide the reuse of validated capabilities and targeted teaching for missing skills.}
    \label{fig:workflow}
  \end{minipage}\par
  \vspace{0.5\baselineskip}
}
\maketitle
\thispagestyle{headings}
\pagestyle{headings}

\begin{abstract}
Executing an unfamiliar laboratory protocol requires a robot to identify missing capabilities and determine whether each operation remains valid as the experiment progresses. This letter presents BioAgent, an embodied agent that addresses these coupled challenges of capability sufficiency and experimental validity.  It matches human-reviewed requirements to validated skills and task packages, selects tool-based implementations for missing behaviors, and requests human teaching when needed during learning. Learned skills undergo sequential task-package validation to check outcomes and compatibility between successive operations. During execution, precondition, outcome, and experimental-state checks govern progression, autonomous recovery, or termination. Execution evidence guides tool selection, while accepted experience supports offline policy updates and task packaging. On the second task, BioAgent reuses 17 skill types unchanged,
revises two existing implementations, and acquires five new types. On two dual-arm laboratory tasks with 20 and 31 ordered skill invocations, BioAgent achieves complete-task success rates of 66\% and 58\% with autonomous recovery over 50 runs per task. Across 25 atomic skills evaluated over 50 autonomous trials each with recovery disabled, success rates range from 84\% to 100\%, with 17 skills achieving 50/50. Project page:{\normalfont 
{\color{projectblue}\urlstyle{same}\url{https://suiuko.github.io/bioagentweb/}}.}
\end{abstract}

\section{Introduction}
A robot prepared for one laboratory protocol is not necessarily ready for the next. New protocols can require operations or skill combinations beyond its finite repertoire. Even when the required skills are available, earlier steps or failures can change the conditions under which they can be used. Executing an unfamiliar protocol therefore poses two coupled challenges: \emph{capability sufficiency}, whether the robot has the capabilities required by the task, and \emph{experimental validity}, whether the current state permits execution under the specified constraints.

Automated laboratory systems carry out routine handling operations as part of multi-step experimental workflows~\cite{miles2018,roboculture,miller2020}. More recent systems use protocol instructions to guide robotic planning~\cite{biomars} and vision--language--action (VLA) models to execute laboratory manipulations~\cite{bioprovla,protoact}. However, translating a protocol into a plan does not ensure that the robot can perform every required operation. Even when most of the workflow is supported by existing skills or their compositions, a single missing behavior can prevent its completion. Guided demonstration and continual skill learning provide ways to acquire additional capabilities~\cite{diehl2025,gu2024}. The challenge is therefore to identify which protocol requirements remain unsupported and direct learning toward those gaps while retaining validated skills for reuse.

Capability sufficiency alone does not establish experimental validity. Each operation acts on the state produced by preceding steps. After a partially completed liquid transfer, for example, the robot may still be able to repeat the motion, but doing so may no longer satisfy the experimental requirements. The decision to continue depends on the current sample, container, and liquid states and the protocol's timing constraints. A missing capability calls for acquisition; an unmet but restorable precondition calls for preparation or recovery; a confirmed validity violation requires termination.

This letter presents BioAgent, an embodied agent system that enables robots to execute unfamiliar laboratory protocols by closing only the capability gaps that prevent valid execution, while retaining validated capabilities for future reuse (Fig.~\ref{fig:workflow}). BioAgent matches human-reviewed requirements to validated skills and task packages, then refines unsupported requirements to distinguish reusable compositions from missing atomic behaviors. For each missing behavior, it selects or combines tools and requests human demonstrations or corrections only when needed during learning. Newly learned skills then undergo sequential task-package validation, which checks both their outcomes and the conditions linking successive operations.

During execution, precondition, outcome, and experimental-state checks determine whether to proceed, retry, restore preconditions, or terminate. A missing recovery skill identified during development returns the affected requirement to learning. Execution evidence guides subsequent tool choices, while accepted demonstrations, execution traces, and corrections support offline policy updates and task packaging. Validated candidates enter the reusable repertoire, linking task execution to capability growth in subsequent experiments.

Our contributions are threefold:
\begin{enumerate}
\item We develop a \emph{targeted capability-acquisition procedure} that matches task requirements to validated skills and compositions, identifies missing behaviors, and directs tool-based skill construction and, when needed, human teaching toward those gaps.
\item We develop a \emph{recovery mechanism that couples experimental validity with capability acquisition}. BioAgent uses observed partial effects to identify admissible recovery, composes validated skills or requests targeted teaching for missing recovery behaviors during learning, and reassesses sample, container, liquid-state, and timing constraints before resumption.
\item We present an \emph{acquisition--execution--reuse mechanism} that separates skill learning from sequential task-package validation, uses execution evidence to guide tool selection, and converts accepted experience into candidate policies and task packages for validation and reuse.
\end{enumerate}

Experiments on two dual-arm laboratory tasks demonstrate BioAgent's ability to close capability gaps selectively while retaining validated skills for reuse. On the second task, BioAgent reuses 17 skill types unchanged, revises two implementations, and acquires five new skill types, expanding the validated repertoire from 20 to 25 types. Validity-aware recovery improves complete-task success by 12 and 18 percentage points on the two tasks, respectively.

\section{Related Work}
\subsection{Agent-Driven Skill Acquisition}
Agentic systems connect skill acquisition with the improvement of existing behaviors. Agentic Skill Discovery (ASD) uses language models to generate rewards and success criteria for reinforcement learning~\cite{asd}, while ENPIRE searches over controllers and learning procedures through real-robot trials~\cite{enpire}. RoboClaw connects data collection, policy learning, and recovery from execution failures~\cite{roboclaw}; ASPIRE retains validated repairs as reusable in-context skills~\cite{aspire}.

During execution of a multi-step laboratory protocol, learning needs arise from the requirements and dependencies of the full workflow. BioAgent ties acquisition to these requirements: it retains validated skills, decomposes unsupported operations, and requests targeted teaching only when additional learning is needed. 

\subsection{Tool and Policy Orchestration}
Tool and policy orchestration focuses on selecting and composing available capabilities. SayCan combines language-based task relevance with learned affordances~\cite{saycan}, while Code as Policies generates programs over perception and control APIs~\cite{codeaspolicies}. WetRobo adapts laboratory programs using observations, demonstrations, and external tools~\cite{wetrobo}. RoboHarness uses capability metadata and execution memory to orchestrate heterogeneous robot policies~\cite{roboharness}.

Beyond selecting existing capabilities, BioAgent uses tool orchestration to construct missing skills. It specifies each skill's required operation, preconditions, and outcome separately from its tool-based implementation. When validated skills or their compositions cannot meet a requirement, BioAgent uses execution experience to select or combine tools for a new or revised implementation, requesting human teaching when needed. New or revised implementations are validated within the task sequence before reuse.

\subsection{Laboratory Protocol Execution and Recovery}
Robotic laboratory systems connect written experimental procedures to physical execution. RoboCulture integrates liquid handling, tip exchange, and culture monitoring through behavior trees~\cite{roboculture}. BioMARS combines experimental-procedure generation with function execution and anomaly detection~\cite{biomars}. ProtoAct translates laboratory protocols into structured subtasks and monitoring conditions for robotic learning~\cite{protoact}, while BioProVLA-Agent incorporates precondition checks, execution verification, retries, and human intervention~\cite{bioprovla}.

Recovery in laboratory protocols must account for changes to samples, containers, and liquids, because repeating a partially completed operation can violate protocol constraints. BioAgent uses the current experimental state to distinguish admissible retries from precondition restoration. Missing recovery capabilities trigger targeted teaching during learning.

\begin{figure*}[t]
\centering
\includegraphics[width=\textwidth]{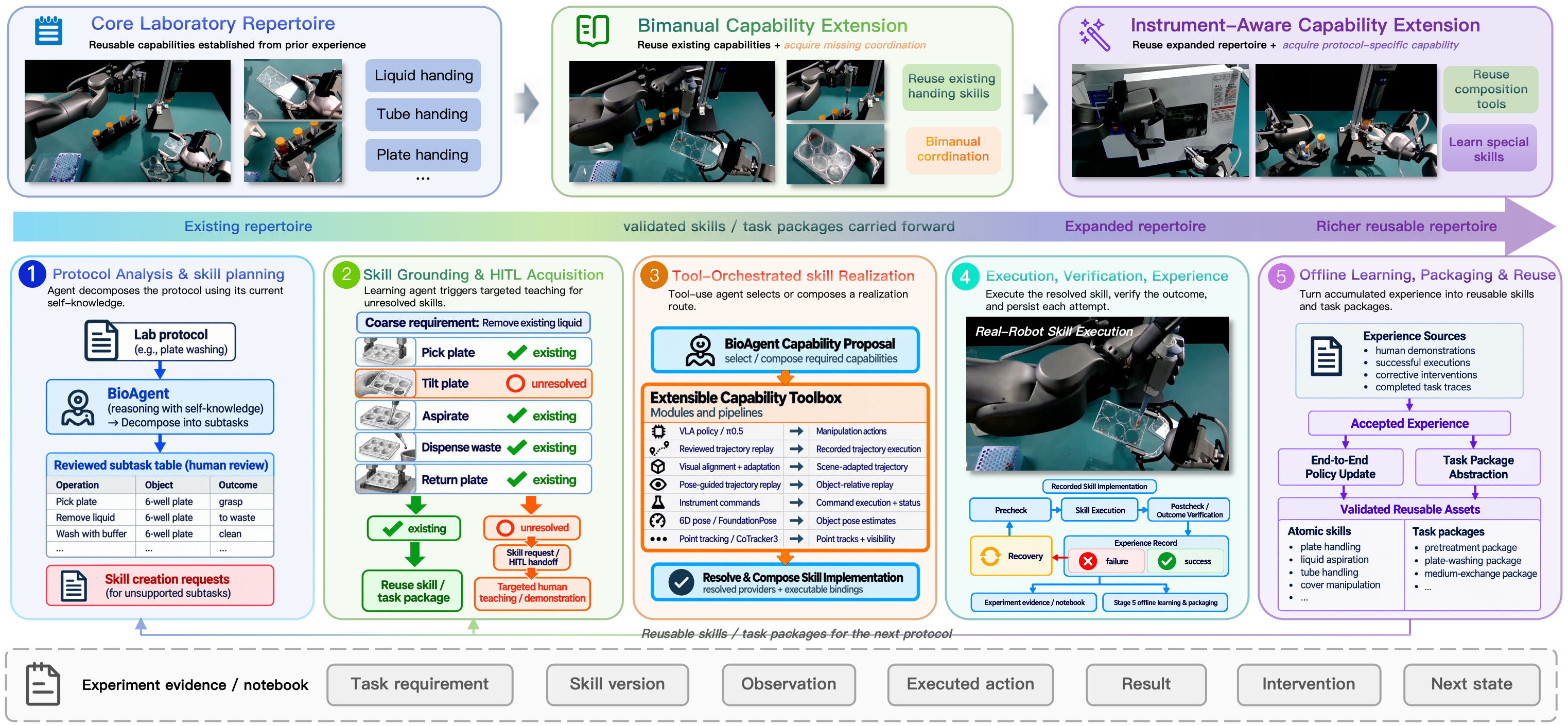}
\caption{BioAgent architecture. Protocol requirements guide capability matching, targeted teaching, and tool-based skill implementation. Verified execution produces experience for offline policy learning and task packaging in Stage~5; validated capabilities return to planning in Stage~1. The upper row illustrates the extension from laboratory handling to bimanual and instrument-related operations. Execution records are retained in the experiment notebook.}
\vspace{-2mm}
\label{fig:architecture}
\end{figure*}

\section{Method}
BioAgent closes capability gaps through targeted learning and sequential task package validation (Fig.~\ref{fig:architecture}; Algorithm~\ref{alg:bioagent}). The Analyzer derives requirements, the Planner acquires  and reuses skills, and the Monitor checks execution and updates tool use experience. The validation process identifies necessary skill and transition revisions.

\subsection{Problem Formulation and Capability Representation}
\label{sec:formulation}
\label{sec:skills}
\subsubsection{Experimental requirements}
Given a laboratory request and protocol instructions $P$, the Analyzer uses its knowledge to form ordered operation--object--outcome requirements:
\begin{equation}
 Q(P)=(q_1,\ldots,q_n),\qquad q_i=(o_i,b_i,g_i).
 \label{eq:requirements}
\end{equation}
Here, $o_i$ denotes an operation, $b_i$ its objects and parameters, and $g_i$ its required outcome. Each requirement specifies preconditions and constraints on samples, containers, liquid states, and timing. Human review confirms their meaning and order. Capability matching refines $Q(P)$ into skill-level requirements. A task is considered successful only if all mandatory outcomes are achieved in the specified order and all constraints are maintained throughout execution.

\subsubsection{Skills, packages, and tools}
BioAgent maintains tools $\T$ and persistent validated skill and task package libraries $\K,\B$ with initial $\K_0=\B_0=\varnothing$. An atomic skill specifies an operation, bindings, preconditions, and a verifiable outcome; its record links these to an implementation, development stage, and validation evidence. Implementations may contain multiple motions or tool calls. Task packages retain versioned skill references, bindings, order, endpoint requirements, and inter-operation constraints.

The toolbox comprises execution backends and pipelines, a device interface, and perception tools (Table~\ref{tab:tools}). Skill implementations use VLA policy inference, reviewed trajectory replay with visual or object pose guided adaptation when needed, or instrument command execution. 

\subsubsection{Learning and validation stages}
\emph{Learning} constructs or revises skills with human teaching when needed. Once all required skills are learned or validated, \emph{validation} checks their ordered composition and registers passing candidates in $\K,\B$.

\begin{table}[!t]
\caption{Tools and tool chains grouped by their primary function.}
\label{tab:tools}
\centering
\footnotesize
\setlength{\tabcolsep}{2.5pt}
\renewcommand{\arraystretch}{1.00}
\begin{tabularx}{\columnwidth}{@{}>{\raggedright\arraybackslash}p{0.28\columnwidth}>{\raggedright\arraybackslash}p{0.15\columnwidth}Y@{}}
\toprule
Module or pipeline & Type & Function in BioAgent \\
\midrule
VLA policy inference & Execution module & Predict manipulation actions from visual observations, robot state, and task instructions using $\pi_{0.5}$~\cite{pi05}. \\
\addlinespace[1pt]
Reviewed trajectory replay & Execution backend & Execute a reviewed demonstration trajectory under its specified starting conditions. \\
\addlinespace[1pt]
Visual alignment and trajectory adaptation & Execution pipeline & Align task-relevant objects with demonstration references and adapt the recorded trajectory to the current scene. \\
\addlinespace[1pt]
Object-pose-guided trajectory replay & Execution pipeline & Use estimated object poses to adapt and execute a reviewed object-relative trajectory. \\
\addlinespace[1pt]
Instrument command execution & Device interface & Execute supported instrument command sequences and return execution status. \\
\addlinespace[1pt]
6D object pose estimation & Perception tool & Estimate the target object's pose in a specified coordinate frame, e.g., using FoundationPose~\cite{foundationpose}. \\
\addlinespace[1pt]
Visual point tracking & Perception tool & Return trajectories and visibility estimates for selected image points, e.g., using CoTracker3~\cite{cotracker3}, as evidence for subsequent checks. \\
\bottomrule
\end{tabularx}
\end{table}

\subsection{Closing the Capability Gap}
\label{sec:acquisition}
\subsubsection{Matching before learning}
The Planner matches requirements to compatible skills and packages in $\K,\B$, storing the assignments in $\mathcal{M}$. It refines unmatched requirements to identify reusable skill compositions and missing behaviors. The remaining capability gaps are defined as $\Delta$, while new or revised implementations enter the candidate set $\mathcal{C}$. Matched skills with unmet preconditions require preparation or recovery. In plate washing (Fig.~\ref{fig:architecture}), a missing tilting behavior triggers targeted teaching while pickup, aspiration, waste dispensing, and return remain assigned to existing skills.

\subsubsection{Experience-guided learning plans}
For skill acquisition or revision, the Planner uses tool-use experience $E_t(u,c)$ to select or combine tools for implementing the required behavior. Here, $E_t(u,c)$ summarizes the observed performance of execution module $u$ for skill category $c$ at monitoring update $t$ (Sec.~\ref{sec:offline}). Given candidate implementations $\mathcal{R}(q)$, their associated tools $\T_q$, and the operation category $c(q)$, the implementation plan is selected as
\begin{equation}
\tau_q=F_{\mathrm{select}}\!\left(
\mathcal{R}(q),
\{E_t(u,c(q))\}_{u\in\T_q}
\right),
\label{eq:select}
\end{equation}
where $\tau_q$ denotes the selected implementation plan. If a selected tool has $E_t(u,c(q))<\eta$, the Planner reassesses the implementation and requests additional teaching during learning when needed.

The plan specifies tools, operation, objects, preconditions, outcome, and teaching needs. 
Teaching data $H_q$ provide demonstrations for policy adaptation or reviewed reference trajectories for replay:
\begin{equation}
 \tilde{s}_q=F_{\mathrm{realize}}(q,\tau_q,H_q),
 \label{eq:realize}
\end{equation}
where $F_{\mathrm{realize}}$ constructs the skill implementation, and $H_q=\varnothing$ indicates that no additional teaching data is required when the available tools are sufficient.

\subsubsection{Transition to validation}
Candidate skills enter sequential task package validation after achieving at least 60\% success in learning-stage trials. Once all requirements are satisfied, the Planner assembles candidate task package $\tilde{b}$ from $Q(P)$, $\mathcal{M}$, and $\mathcal{C}$.

\begin{algorithm}[!t]
\caption{Capability-gap closure by learning and sequential validation}
\label{alg:bioagent}
\footnotesize
\begin{algorithmic}[1]
\Require Reviewed requirements $Q(P)$; $\K,\B$; tools $\T$; experience $E$
\Ensure Validated updates, records, and unresolved requirements
\State Match and refine $Q(P)$; retain $\mathcal{M}$ and identify $\Delta$
\State $\mathcal{C}\gets\varnothing$
\Repeat
    \For{each $q\in\Delta$ requiring skill acquisition or revision}
        \State Select $\tau_q$ by Eq.~\eqref{eq:select}; request $H_q$ only if needed
        \State Construct or revise $\tilde{s}_q$ by Eq.~\eqref{eq:realize}; practice it
        \State Retain $\tilde{s}_q$ and its learning status in $\mathcal{C}$
    \EndFor
    \If{any required skill is neither learned nor validated}
        \State \Return $\K,\B$, records, and unresolved requirements
    \EndIf
    \State Assemble or revise $\tilde{b}$ using $Q(P)$, $\mathcal{M}$, and $\mathcal{C}$
    \State $V\gets$ sequential package validation (Sec.~\ref{sec:execution})
    \If{the package and candidates pass the applicable checks}
        \State Register passing skills and $\tilde{b}$ in $\K,\B$ with their scope
        \State \Return $\K,\B$ and records; no unresolved requirements
    \EndIf
    \State $\Delta\gets$ requirements needing revision according to monitor
    \State Refine and rematch $\Delta$; retain unaffected $\mathcal{M},\mathcal{C}$
\Until{development stops or no supported revision remains}
\State \Return $\K,\B$, records, and unresolved requirements
\end{algorithmic}
\end{algorithm}

\subsection{Validity-Aware Execution and Recovery}
\label{sec:execution}
\subsubsection{Sequential validation and targeted revision}
Sequential validation executes $\tilde{b}$, checking outcomes and compatibility with subsequent preconditions and experimental constraints. The record monitor collects outcomes, transitions, partial effects, recovery needs, and exit reasons. Algorithm~\ref{alg:bioagent} summarizes the learning--validation loop.

Evidence monitor directs skill revision for inadequate behavior or tool use, binding or composition revision for transition mismatches, and acquisition for skills missing from admissible recovery plans. The Planner updates and rematches $\Delta$, retains unaffected entries in $\mathcal{M},\mathcal{C}$, and revalidates the package.

\subsubsection{Execution checks}
The Monitor checks skill and task preconditions, safety, and experimental constraints before each operation, then verifies its outcome and the next operation's applicability (Stage~4 in Fig.~\ref{fig:architecture}). The Monitor permits progression only after verifying the current outcome and confirming that the next operation is admissible. It accounts for partial effects when evaluating the resulting state and terminates the run if a safety or experimental-validity violation is confirmed. If uncertainty persists after bounded re-observation, it requests clarification during learning or terminates the run during evaluation.

\subsubsection{Recovery and resumption}
After a failure, the Monitor assesses likely causes and partial effects from observations, actions, and outcome checks. The Planner selects an appropriate recovery strategy by assessing whether the operation can be retried, its preconditions can be restored, and the necessary skills are available.

\emph{R1 (retry)}, BioAgent retries the operation only if its preconditions remain satisfied, the observed partial effects allow another attempt, and the retry budget has not been exhausted. 
\emph{R2 (precondition restoration)}, it composes validated skills to restore unmet preconditions while preserving experimental constraints.

\emph{R3 (missing recovery capability)} applies when admissible recovery requires a behavior unavailable in the validated skill library or through compositions of existing skills. During learning, the Planner adds the missing requirement to $\Delta$, specifies the failure context, restoration target, and constraints, and initiates skill acquisition (Sec.~\ref{sec:acquisition}). Once the acquired skill meets the learning criterion, the Planner incorporates it into a candidate task package for sequential validation. Failed validation returns the affected requirements to learning.

Recovery records guide subsequent implementation selection and skill revision. Before resumption, the Monitor verifies recovery outcomes, continuation preconditions, and experimental constraints on samples, containers, liquids, and timing. During evaluation, no new recovery skills are acquired. The Monitor terminates execution when no admissible continuation exists, including when a required recovery capability is unavailable, or when the recovery budget is exhausted.

\begin{figure*}[t]
\centering
\clipbox{0 {.16\height} 0 {.16\height}}{\includegraphics[width=\textwidth]{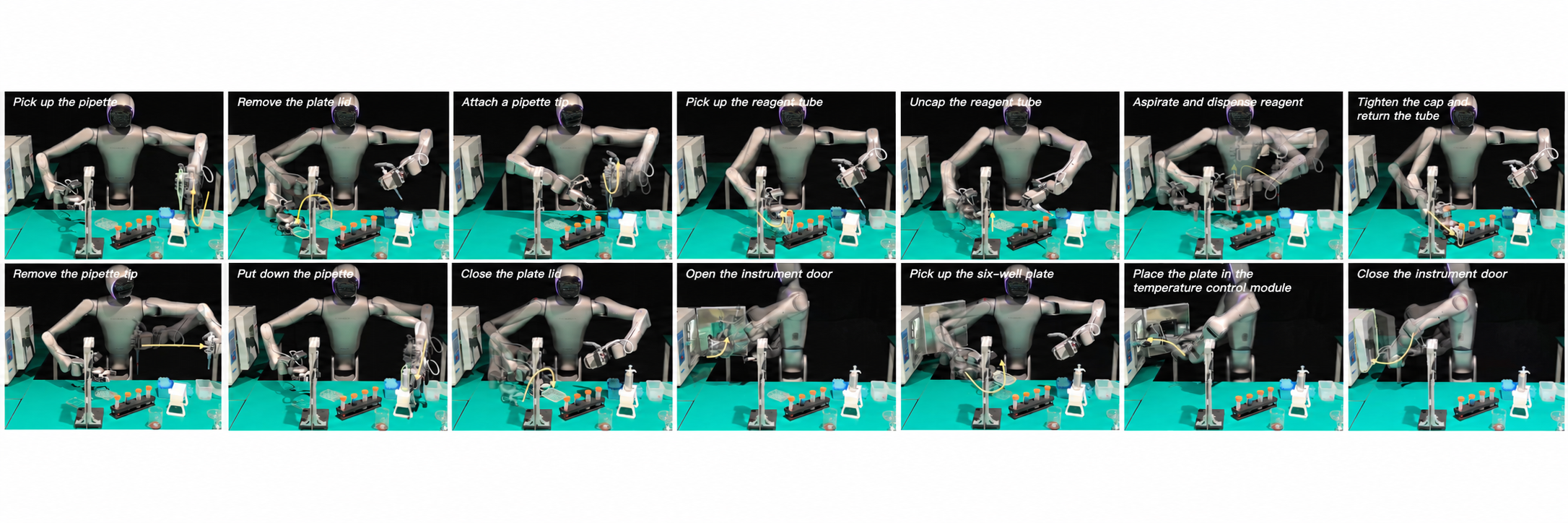}}
\caption{Task~1: plate pretreatment and incubator loading on the Monte~02 dual-arm platform. Representative frames show pipette and reagent-tube handling, pretreatment-liquid transfer, and incubator loading, arranged from left to right and top to bottom.}
\label{fig:task1}
\end{figure*}

\begin{figure*}[t]
\centering
\includegraphics[width=\textwidth]{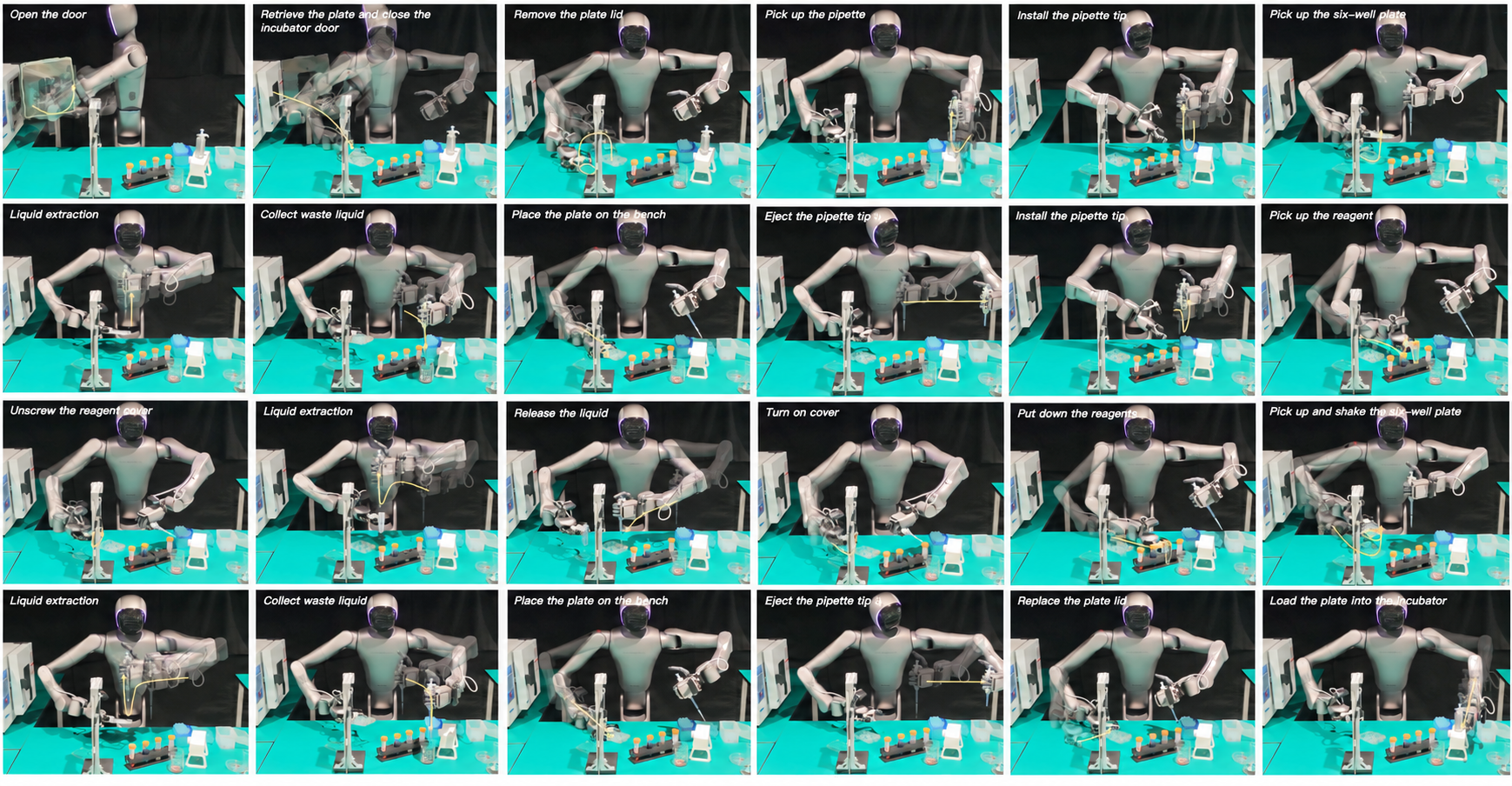}
\caption{Task~2: plate washing. Representative frames show incubator retrieval, removal of existing liquid, cleaning-liquid transfer, and wash-liquid removal, interleaved with tube handling and capping-machine operations. Frames are arranged from left to right and top to bottom.}
\label{fig:task2}
\end{figure*}

\subsection{Consolidation and Growth of Capability}
\label{sec:offline}
Online experience guides tool selection; offline learning updates policies or task compositions, producing candidates for learning and validation.

\subsubsection{Execution records and tool experience}
The notebook retains requirements, skill versions, implementations, observations, actions, outcomes, interventions, successor states, and run exit reasons. During learning and validation, tool-attributable evidence $\epsilon_t^{u,c}$ retains context, verified outcome, and intervention labels; at update $t$,
\begin{equation}
 E_{t+1}(u,c)=F_{\mathrm{exp}}\bigl(E_t(u,c),\epsilon_t^{u,c}\bigr),
 \label{eq:experience}
\end{equation}
where $F_{\mathrm{exp}}$ updates entries supported by tool-specific evidence and retains others. Aggregate outcomes remain linked to the full implementation. Experience guides Eq.~\eqref{eq:select}; policy updates occur offline.

\subsubsection{Offline Learning with Accepted/Validated Experience}
Stage~5 (Fig.~\ref{fig:architecture}) collects verified demonstration, execution, and correction segments in $\D_{\mathrm{acc}}$ with task and intervention context for policy learning and package construction. Failed and unverified attempts are recorded separately in the experiment notebook for diagnosis, while module-attributable evidence is used to update execution module experience.

Compatible observations, actions, and task conditioning yield policy-training data $\D_\pi$ for $\pi_\theta$; ordered accepted traces form $\Hacc$. They produce
\begin{equation}
 \begin{aligned}
 \tilde{\theta}&=F_{\mathrm{learn}}(\theta,\D_\pi),\\
 \tilde{\B}&=F_{\mathrm{package}}(\Hacc;\K).
 \end{aligned}
 \label{eq:offline}
\end{equation}
Verified trajectories generated by visual servoing or VLA policies such as $\pi_{0.5}$ can be incorporated into $\D_\pi$ as supervised training data for subsequent policy updates. Meanwhile, task packages retain versioned skill references, bindings, execution order, and experimental constraints (Sec.~\ref{sec:formulation}), enabling changes to skill composition without retraining the underlying atomic policies.

\subsubsection{Validation and reuse}
Updated policies enter sequential validation after meeting the learning criterion; package validation checks inter-skill transitions. Passing candidates are registered with scope, version, and validation evidence; others return for revision. Later protocols reuse compatible entries and address remaining gaps.

\begin{table}[!t]
\vspace{-2mm}
\caption{Task scope and complete-task success with and without recovery (50 runs per task and condition).}
\label{tab:tasks}

\centering
\footnotesize
\setlength{\tabcolsep}{2pt}
\renewcommand{\arraystretch}{1.00}
\begin{tabular*}{\columnwidth}{@{\extracolsep{\fill}}lccccc@{}}
\toprule
 & & & & \multicolumn{2}{c}{Complete-task count (SR)} \\
\cmidrule(l){5-6}
Task & \shortstack{Ordered\\invocations} & \shortstack{Distinct\\skills} & \shortstack{Runs per\\condition} & \shortstack{Without\\recovery} & \shortstack{With\\recovery} \\
\midrule
Task~1 & 20 & 20 & 50 & 27/50 (54\%) & 33/50 (66\%) \\
Task~2 & 31 & 24 & 50 & 20/50 (40\%) & 29/50 (58\%) \\
\bottomrule
\end{tabular*}
\vspace{-2mm}
\end{table}

\begin{table*}[!t]
\caption{Atomic-skill success rates (\%) over 50 real-robot trials per method and skill on the Monte 02 platform, with recovery disabled.}
\label{tab:skills}
\centering
\small
\setlength{\tabcolsep}{5.4pt}
\renewcommand{\arraystretch}{1.00}
\begin{tabularx}{\textwidth}{@{}rYcccc@{}}
\toprule
 & & \multicolumn{4}{c}{Success rate (\%): 50 trials per method and skill} \\
\cmidrule(l){3-6}
No. & Atomic skill / operation binding & \shortstack{ProtoAct~\cite{protoact}\\(SmolVLA)} & \shortstack{BioProVLA~\cite{bioprovla}\\(AugSmolVLA)} & ENPIRE~\cite{enpire} & BioAgent \\
\midrule
\multicolumn{6}{@{}l}{\textit{Plate handling}} \\
1 & Pick up the six-well culture plate from the target bench position & 42 & 66 & 50 & \best{100} \\
2 & Retrieve the six-well culture plate from the tray inside the incubator & 38 & 68 & 48 & \best{94} \\
3 & Place the six-well culture plate at the target position on the bench & 70 & 82 & 66 & \best{100} \\
4 & Place the six-well culture plate onto the tray inside the incubator & 36 & 56 & 46 & \best{100} \\
5 & Pick up the six-well culture plate lid from the target resting position & 68 & 76 & 54 & \best{100} \\
6 & Set the six-well culture plate lid aside within the laboratory work area & 40 & 62 & 48 & \best{100} \\
7 & Remove the lid from the six-well culture plate on the laboratory bench & 38 & 56 & 50 & \best{100} \\
8 & Replace the lid on the six-well culture plate & 56 & 30 & 24 & \best{86} \\
9 & Mix the liquid contents of the six-well culture plate by shaking & 34 & 46 & 42 & \best{100} \\
\addlinespace[1pt]
\multicolumn{6}{@{}l}{\textit{Pipetting}} \\
10 & Pick up the laboratory pipette from the target resting position & 42 & 62 & 52 & \best{100} \\
11 & Return the laboratory pipette to the target storage position & 44 & 60 & 52 & \best{100} \\
12 & Attach a disposable pipette tip from the tip box to the pipette & 20 & 32 & 46 & \best{88} \\
13 & Eject the attached disposable pipette tip from the laboratory pipette & 36 & 56 & 54 & \best{100} \\
14 & Aspirate liquid reagent from the centrifuge tube using the pipette & 40 & 58 & 46 & \best{100} \\
15 & Aspirate liquid from the six-well culture plate using the pipette & 42 & 62 & 44 & \best{96} \\
16 & Dispense liquid from the pipette into the six-well culture plate & 54 & 68 & 60 & \best{98} \\
17 & Dispense liquid from the pipette into the designated waste container & 32 & 52 & 42 & \best{100} \\
\addlinespace[1pt]
\multicolumn{6}{@{}l}{\textit{Tube handling and capping-machine operations}} \\
18 & Pick up the centrifuge tube from the target resting position & 44 & 60 & 52 & \best{100} \\
19 & Return the centrifuge tube to its original position in the workspace & 48 & 58 & 48 & \best{92} \\
20 & Position the centrifuge tube in the capping machine for uncapping & 46 & 64 & 54 & \best{100} \\
21 & Position the centrifuge tube in the capping machine for recapping & 36 & 40 & 46 & \best{84} \\
22 & Uncap the centrifuge tube using the laboratory capping machine & 56 & 70 & 58 & \best{100} \\
23 & Tighten the centrifuge tube cap using the laboratory capping machine & 42 & 38 & 52 & \best{84} \\
\addlinespace[1pt]
\multicolumn{6}{@{}l}{\textit{Instrument access}} \\
24 & Open the incubator door & 12 & 18 & 22 & \best{100} \\
25 & Close the incubator door & 14 & 20 & 34 & \best{100} \\
\bottomrule
\end{tabularx}
\end{table*}

\begin{samepage}
\section{Experiments}
We evaluate capability acquisition and reuse (RQ1), the effect of validity-aware recovery on task completion (RQ2), and differences in reliability across tasks and operation types (RQ3).\par
\end{samepage}

\subsection{Experimental Platform and Policy Training}
\begin{figure}[!t]
    \centering
    \includegraphics[width=\columnwidth]{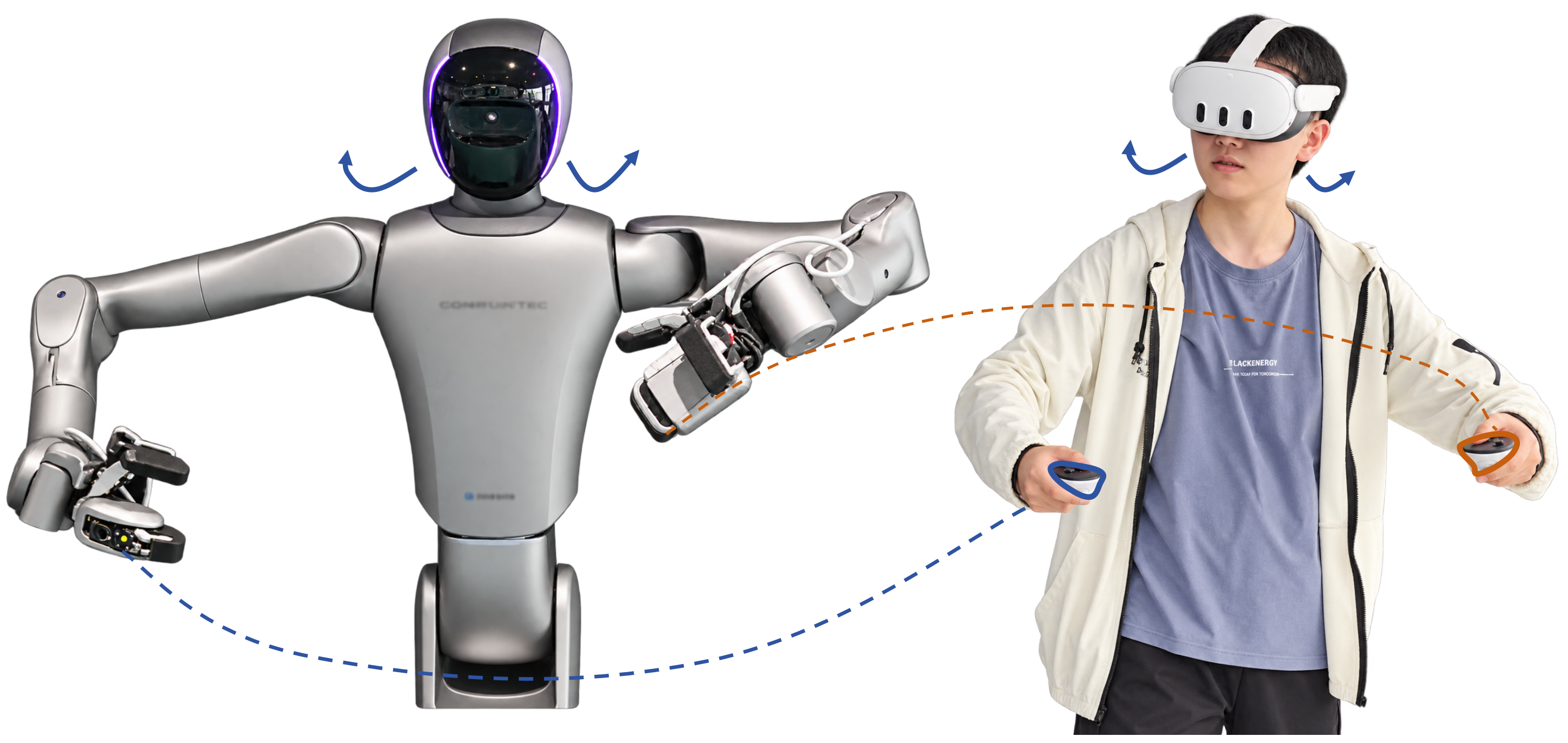}
    \caption{VR-based teleoperation setup for collecting human demonstrations for robotic skill acquisition.}
    \label{fig:vr}
\end{figure}
We use a Monte~02 dual-arm robot from Corenetic AI~\cite{corenetic}, with VR and 3D-mouse teleoperation to collect demonstrations (Fig.~\ref{fig:vr}). We fine-tune $\pi_{0.5}$ from \texttt{pi05\_base} using eight NVIDIA H100 GPUs and perform inference on a local NVIDIA 5090 GPU with 32\,GB of memory. We use task-derived language prompts, a batch size of 32, 30,001 optimization steps, a 16-step action horizon, and a nominal dataset rate of 15\,Hz, while adopting AdamW with gradient clipping at 1.0 and an exponential moving average with a decay of 0.999. We increase the learning rate to $5\times10^{-5}$ over 10,000 warmup steps and keep it constant thereafter.

\subsection{Tasks and Evaluation Metrics}
Task~1 combines lid handling, pipetting, tube uncapping and recapping, and instrument access to transfer pretreatment liquid into a six-well plate and load the plate into the incubator (Fig.~\ref{fig:task1}). Task~2 retrieves the plate, removes existing liquid, adds cleaning liquid, mixes the contents, and removes wash liquid, interleaving tip exchange, tube handling, and labware return (Fig.~\ref{fig:task2}). The tasks contain 20 and 31 ordered invocations of 20 and 24 distinct skill types, respectively (Table~\ref{tab:tasks}). Skill types are defined by operations and object or destination bindings. Repeated invocations count separately toward sequence length but once toward the distinct-skill count. Reagent identity parameterizes tube aspiration; tube positioning and machine-driven cap tightening are separate skills.

Table~\ref{tab:tasks} reports 50 runs per task and condition; Table~\ref{tab:skills} reports 50 trials per method--skill pair. The atomic-skill success rate is
\begin{equation}
 \SR(m,s)=\frac{k_{m,s}}{50}\times100\%,
 \label{eq:sr}
\end{equation}
where $k_{m,s}$ counts successful trials for method $m$ and skill $s$. A complete-task run succeeds when all mandatory outcomes are achieved in order, protocol constraints are maintained throughout execution, and the final state passes all required checks.

\subsection{Comparison Conditions}
\label{sec:comparison}
We adapt the ProtoAct (SmolVLA)~\cite{protoact}, BioProVLA (AugSmolVLA)~\cite{bioprovla}, and ENPIRE~\cite{enpire} frameworks to our Monte~02 laboratory setup. None of the adapted baselines completes the full 20-invocation workflow in our Task~1 settings. Table~\ref{tab:skills} therefore compares the methods at the atomic-skill level, using 50 trials per method--skill pair with recovery disabled. All methods are evaluated with the same objects, initial conditions, and skill-specific success criteria. ProtoAct, BioProVLA, and BioAgent use training data from the same source and matched training budgets; ENPIRE undergoes platform adaptation without additional model training.

For BioAgent, Table~\ref{tab:tasks} compares complete-task execution without recovery and with R1+R2 over 50 runs per task and condition. A separate experiment compares no recovery, R1 only, and R1+R2 over 40 runs per task and configuration; Table~\ref{tab:prefix} reports progress and final outcomes. All evaluation runs are autonomous. BioAgent's recovery follows the execution checks in Sec.~\ref{sec:execution}.

\begin{figure}[t]
\centering
\includegraphics[width=\columnwidth]{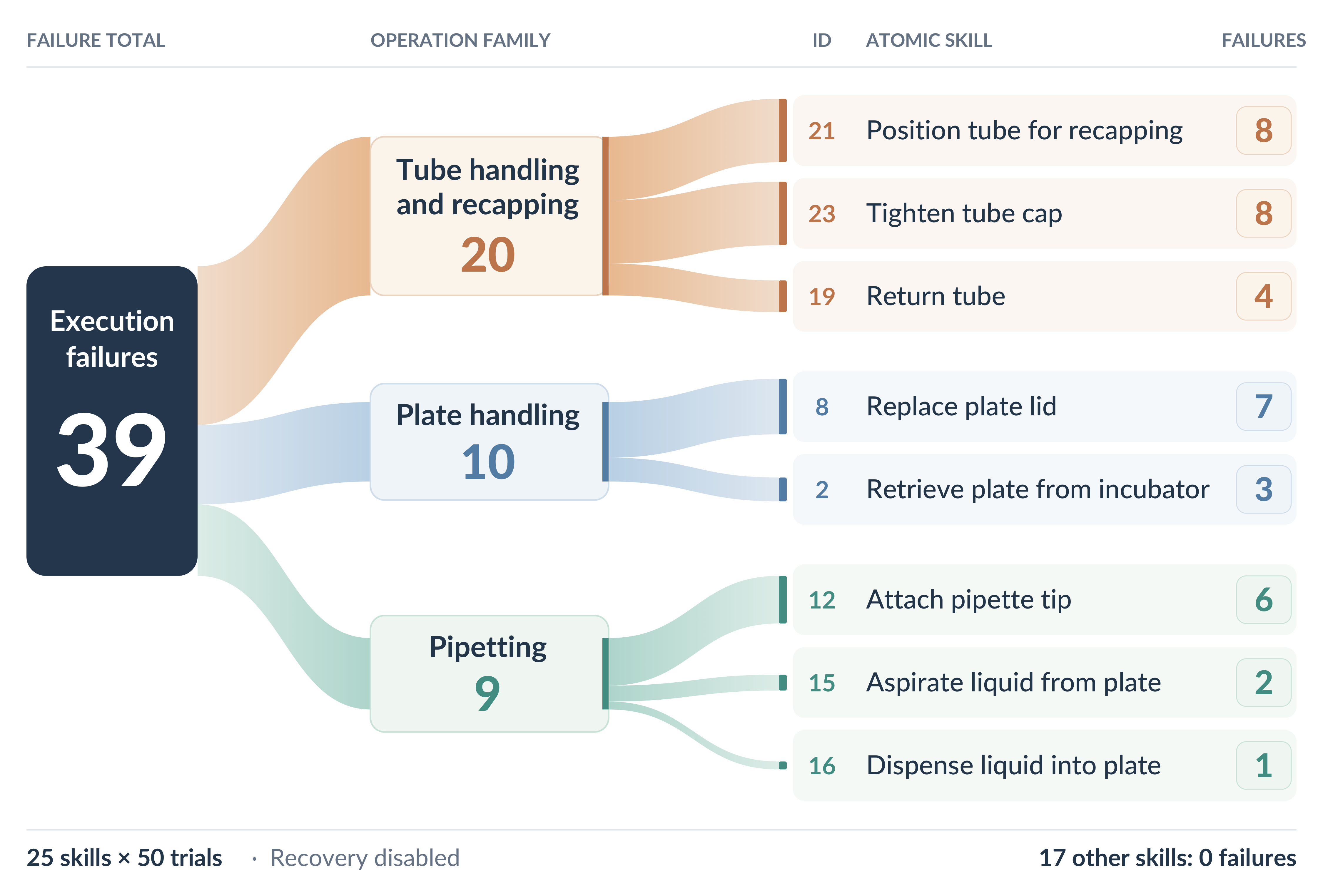}
\caption{Distribution of the 39 failures in BioAgent's atomic-skill evaluation (Table~\ref{tab:skills}), with 50 trials per skill and recovery disabled. The other 17 skills succeed in all 50 trials.}
\vspace{-2mm}
\label{fig:failure-profile}
\end{figure}

\subsection{RQ1: Can BioAgent Close Capability Gaps and Reuse Skills?}
\label{sec:rq1}
BioAgent closes all confirmed capability gaps while retaining validated skills (Table~\ref{tab:acquisition}). The workflows share 19 of their 25 distinct skill types, covering 19/24 (79.2\%) of Task~2's requirements.

Starting from an empty library, BioAgent acquires 20 validated skill types for Task~1 and closes all 20 initial gaps. For Task~2, it reuses 17 types unchanged (17/24, 70.8\%), revises two implementations, and acquires five new skill types. All seven flagged gaps are confirmed and closed, expanding the library from 20 to 25 types. Acquisition or revision therefore affects only seven of the 24 required implementations. The difference between 19 shared types and 17 unchanged implementations shows that familiar operations can still require adaptation. Capability matching preserves compatible implementations while targeting the remaining gaps.

Task~2 requires more skill types yet records nine teaching requests and 114 minutes of human participation, compared with 26 requests and 360 minutes for Task~1. It combines substantial reuse with lower recorded teaching effort. Its task package undergoes three validation rounds, versus two for Task~1. Thus selective acquisition limits implementation changes, while sequential validation checks compatibility among reused, revised, and newly acquired skills.

\begin{table}[!t]
\caption{Skill acquisition and reuse during task development.}
\label{tab:acquisition}
\centering
\footnotesize
\setlength{\tabcolsep}{4pt}
\renewcommand{\arraystretch}{1.00}
\begin{tabularx}{\columnwidth}{@{}Ycc@{}}
\toprule
Development quantity & Task~1 & Task~2 \\
\midrule
Initial validated skill types & 0 & 20 \\
Required distinct types & 20 & 24 \\
Planner gap flags & 20 & 7 \\
Reference-confirmed implementation gaps & 20 & 7 \\
Unchanged reused types & 0 & 17 \\
Revised existing types & 0 & 2 \\
New types acquired & 20 & 5 \\
Initial gaps closed / confirmed & 20/20 & 7/7 \\
Teaching requests & 26 & 9 \\
Accepted demonstrations & 180 & 54 \\
Accepted corrections & 26 & 13 \\
Human participation (min) & 360 & 114 \\
Package-validation rounds & 2 & 3 \\
Final primary-task library (types) & 20 & 25 \\
\bottomrule
\end{tabularx}
\end{table}

\begin{table}[!t]
\caption{Task progress and complete-task outcomes in the separate recovery comparison, with 40 runs per task and configuration. Columns indexed by $k$ count \# of runs completing the first $k$ skill invocations; Final counts complete-task successes after terminal checks. All entries are counts out of 40.}
\label{tab:prefix}
\centering
\footnotesize
\setlength{\tabcolsep}{3pt}
\renewcommand{\arraystretch}{1.00}
\begin{tabularx}{\columnwidth}{@{}Yccccc@{}}
\toprule
\multicolumn{6}{@{}l}{\emph{Task~1: 20 invocations}} \\
Condition & $k{=}5$ & 10 & 15 & 20 & Final \\
\midrule
No recovery & 38 & 34 & 27 & 21 & 20 \\
R1 only & 38 & 35 & 29 & 24 & 22 \\
R1+R2 & 39 & 37 & 32 & 27 & 26 \\
\midrule
\multicolumn{6}{@{}l}{\emph{Task~2: 31 invocations}} \\
Condition & $k{=}8$ & 16 & 24 & 31 & Final \\
\midrule
No recovery & 35 & 27 & 20 & 16 & 14 \\
R1 only & 37 & 32 & 25 & 20 & 19 \\
R1+R2 & 37 & 34 & 30 & 24 & 21 \\
\bottomrule
\end{tabularx}
\end{table}

\subsection{RQ2: Does Validity-Aware Recovery Improve Task Completion?}
\label{sec:rq2}
Enabling R1+R2 increases complete-task success by 12 and 18 percentage points for Tasks~1 and~2, respectively (Table~\ref{tab:tasks}). Recovery therefore improves execution after capability development, when the required skills are already available.

In the separate 40-run comparison (Table~\ref{tab:prefix}, \emph{Final}), R1 increases success by 5 and 12.5 percentage points for Tasks~1 and~2, respectively; adding R2 yields further gains of 10 and 5 points. Restoration provides the larger observed increment for Task~1, whereas retries provide the larger increment for Task~2. Both tasks benefit from restoration beyond retries alone.

The tube-recapping operations in Table~\ref{tab:skills} illustrate why retry and restoration are complementary. If cap tightening fails while the tube remains correctly positioned and checks permit another attempt, R1 can repeat the tightening operation. If the tube is misaligned in the capping machine, R2 can first invoke the validated tube-positioning skill before tightening resumes. The observed tube state therefore determines whether tightening can be retried directly or requires repositioning first.

\subsection{RQ3: How Does Reliability Vary Across Tasks and Operations?}
\label{sec:rq3}
Task~2 has 31 skill invocations and 24 distinct types, compared with 20 invocations and 20 types for Task~1. Its sequence is 55\% longer and includes seven repeat invocations. Recovery narrows the success rate gap between tasks from 14 to 8 percentage points (Table~\ref{tab:tasks}), while Task~2 retains a lower completion rate. The comparison reflects both length and operation composition: shared skills supply capabilities, while successful execution also requires compatible transitions.

Table~\ref{tab:prefix} shows that R1+R2 yields higher completion counts than no recovery at every reported checkpoint. For Task~2, 30 of 40 runs complete the first 24 skill invocations with R1+R2, compared with 20 of 40 without recovery. Under R1+R2, 27 Task~1 runs complete all 20 invocations, but one of these fails the terminal checks, leaving 26 complete-task successes. For Task~2, 24 runs complete all 31 invocations, but three of these fail the terminal checks, leaving 21 complete-task successes.

At the atomic-skill level, BioAgent outperforms all three baselines on each of the 25 skills (Table~\ref{tab:skills}), averaging 96.9\% success compared with 41.2\% for ProtoAct, 54.4\% for BioProVLA, and 47.6\% for ENPIRE. This recovery-disabled comparison directly evaluates the skill implementations. BioAgent records 39 failures in 1,250 trials; tube positioning for recapping, cap tightening, plate-lid replacement, tip attachment, and tube return account for 33 (84.6\%; Fig.~\ref{fig:failure-profile}). This concentration identifies specific implementations for refinement while retaining the more reliable skills.

At the complete-task level, 14 Task~1 runs and 19 Task~2 runs fail under R1+R2 (Table~\ref{tab:prefix}). Each receives one primary terminal-failure category: alignment/contact (16), liquid-state (9), transition (4), or timing (4). Alignment/contact and liquid-state failures account for 25/33 cases (75.8\%), while transition and timing failures highlight requirements across skill boundaries. These results identify complementary priorities for BioAgent's acquisition, validation, and recovery mechanisms: refining error-prone implementations and maintaining valid experimental conditions throughout the workflow.

\section{Conclusion}

This letter presented BioAgent for the acquisition and reuse of targeted skills in long-horizon laboratory protocols. BioAgent matches protocol requirements to validated capabilities, acquires missing behaviors, and selects recovery according to the current experimental state. Sequential task-package validation connects skill learning with reuse, while execution experience guides tool selection and accepted experience supports offline policy learning and task packaging. On two dual-arm tasks with 20 and 31 ordered skill invocations, BioAgent achieved complete-task success rates of 66\% and 58\% with autonomous recovery over 50 runs per task. Future work will extend the evaluation of teaching effort, cross-protocol skill reuse, and recovery across a broader range of laboratory protocols.

\end{document}